\documentclass[10pt, conference]{IEEEtran}
\usepackage{lscape}
\usepackage{array}
\usepackage{graphicx}
\usepackage{multicol}
\usepackage{multirow}
\usepackage{amssymb}
\usepackage{pifont}
\usepackage{tikz}
\usepackage{verbatim}
\usepackage{array}
\usepackage{longtable}
\usepackage{supertabular}
\usepackage{amsmath}
\usepackage{epstopdf}
\usepackage{verse}
\usepackage{cite}
\usepackage{color}
\usepackage{lettrine}
\usepackage{hyperref}
\usepackage{soul}
\usepackage{comment}
\usepackage{verbatim}
\usepackage{subcaption}
\usepackage{amsfonts}
\usepackage{multirow}
\usepackage{tabularray}
\usepackage[normalem]{ulem}
\usepackage{tikz}
\usepackage{amsmath}
\usepackage{pifont}
\DeclareTextFontCommand{\emph}{\em}
\DeclareMathOperator{\sign}{sign}

\DeclareMathOperator*{\argmin}{argmin}
\DeclareMathOperator{\EX}{\mathbb{E}}
\DeclareMathOperator{\Rd}{\mathbb{R}}
\usepackage{algorithm}
\usepackage{algorithmic}
\def\BibTeX{{\rm B\kern-.05em{\sc i\kern-.025em b}\kern-.08em
    T\kern-.1667em\lower.7ex\hbox{E}\kern-.125emX}}
\usepackage{placeins}
\begin{document}

\title{
Improving Machine Learning Robustness via Adversarial Training
} 


\author{
\IEEEauthorblockN{Long Dang}
\IEEEauthorblockA{\textit{\textit{ICNS Lab and Cyber Florida}} \\
\textit{University of South Florida}\\
Tampa, FL. USA \\
longdang@usf.edu}

\and 

\IEEEauthorblockN{Thushari Hapuarachchi}
\IEEEauthorblockA{\textit{\textit{ICNS Lab and Cyber Florida}} \\
\textit{University of South Florida}\\
Tampa, FL. USA \\
saumya2@usf.edu}

\and
\IEEEauthorblockN{Kaiqi Xiong}
\IEEEauthorblockA{\textit{\textit{ICNS Lab and Cyber Florida}} \\
\textit{University of South Florida}\\
Tampa, FL. USA \\
xiongk@usf.edu}

\and
\IEEEauthorblockN{Jing Lin}
\IEEEauthorblockA{\textit{\textit{ICNS Lab and Cyber Florida}} \\
\textit{University of South Florida}\\
Tampa, FL. USA \\
jinglin@usf.edu}
}

\maketitle
\begin{abstract}
As Machine Learning (ML) is increasingly used in solving various tasks in real-world applications, it is crucial to ensure that ML algorithms are robust to any potential worst-case noises, adversarial attacks, and highly unusual situations when they are designed. Studying ML robustness will significantly help in the design of ML algorithms. In this paper, we investigate ML robustness using adversarial training in centralized and decentralized environments, where ML training and testing are conducted in one or multiple computers. In the centralized environment, we achieve a test accuracy of 65.41\% and 83.0\% when classifying adversarial examples generated by Fast Gradient Sign Method and DeepFool, respectively. Comparing to existing studies, these results demonstrate an improvement of 18.41\% for FGSM and 47\% for DeepFool. In the decentralized environment, we study Federated learning (FL) robustness by using adversarial training with independent and identically distributed (IID) and non-IID data, respectively, where CIFAR-10 is used in this research. In the IID data case, our experimental results demonstrate that we can achieve such a robust accuracy that it is comparable to the one obtained in the centralized environment. Moreover, in the non-IID data case, the natural accuracy drops from 66.23\% to 57.82\%, and the robust accuracy decreases by 25\% and 23.4\% in C\&W and Projected Gradient Descent (PGD) attacks, compared to the IID data case, respectively. We further propose an IID data-sharing approach, which allows for increasing the natural accuracy to 85.04\% and the robust accuracy from 57\% to 72\% in C\&W attacks and from 59\% to 67\% in PGD attacks.
\end{abstract}


\section{Introduction} \label{introduction}
Machine Learning (ML) is increasingly used in solving various tasks in real-world applications, for example, image and speech recognition, traffic engineering and securing in-vehicle networks~\cite{Li2020AnNetworks}, malware detection~\cite{DBLP:journals/access/LiXCH19} and security attack detection, identification of diseases and diagnosis, as well as drug discovery and manufacturing.
It is crucial to ensure that ML algorithms are robust to any potential worst-case noises, adversarial attacks, and highly unusual situations when they are designed. The studies of ML robustness will greatly help in the design of ML algorithms to reduce or avoid the potentially risky use of ML in various real-world applications.

\begin{figure*}
         \centering
         \includegraphics[scale=0.78]{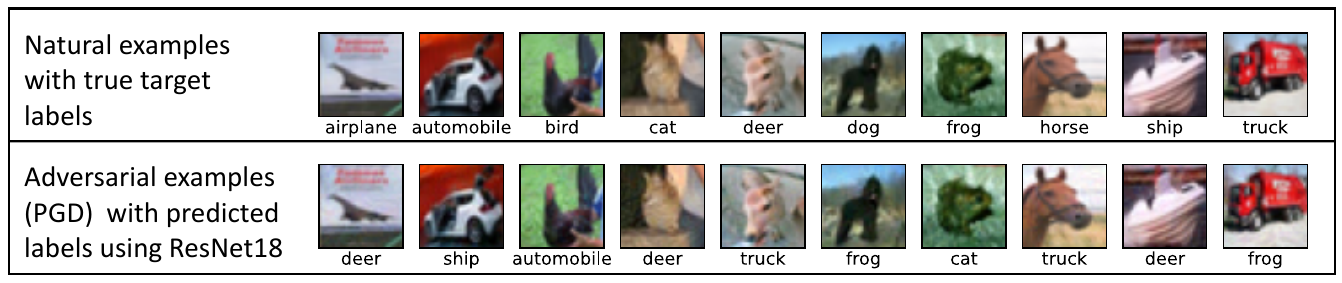}
        \caption{Illustration of adversarial CIFAR-10 images. The predicted labels were obtained based on ResNet18 \cite{He2015ResNet}, and the adversarial examples were generated using the PGD attack with the perturbation of $\epsilon=\frac{8}{255}$.}
        \label{fig:adv_img}
\end{figure*}

In this research, we investigate ML robustness by using adversarial training in the centralized and decentralized environments, where ML training and testing are conducted in one or multiple computers, respectively. In the centralized environment, adversarial attacks (i.e., adversarial examples are introduced in the testing time, or minor noise is intentionally added to testing data) are to mislead a classifier's prediction. Fig.~\ref{fig:adv_img} illustrates how adversarial examples are difficult to be distinguished from natural images by an ML model, but they are easily recognized by human eyes~\cite{DBLP:journals/ejisec/LinNX22}. 
Many defense strategies against adversarial examples have been proposed over the years. Among them, adversarial training is one of the most promising defense techniques against evasion attacks (or adversarial examples)~\cite{DBLP:conf/nips/AndriushchenkoF20, zizzo2020fat, 10.5555/AAI29062992}. 
Lin, et al.~\cite{DBLP:journals/ejisec/LinNX22} introduced soft labeling in adversarial training to against adversarial examples at the test time, where they improved the robust accuracy. 
These results demonstrate an
improvement of 18.41\% for FGSM and 47\% for DeepFool, compared to existing studies in Lin, et al.~\cite{DBLP:journals/ejisec/LinNX22}.

Furthermore, in the decentralized environment, our study focuses on Federated learning (FL) robustness through adversarial training with independent and identically distributed (IID) and non-IID data, respectively, where CIFAR-10 is used in this research.
FL is considered one of the most well-known distributed ML paradigms, and it has many promising applications in the real world. FL leverages up to millions of edge devices, e.g., mobile devices (referred to as clients) to train ML models in their own data locally, and the local model parameters are aggregated through a server (also called an aggregator or a model aggregator)~\cite{DBLP:conf/sp/NasrSH19, DBLP:journals/corr/abs-2106-10196}. FL also alleviates the privacy concerns of local data, so it has been considered an ideal paradigm for the analysis of large sensitive and private data in healthcare, financial, and military applications. However, existing studies showed that robust accuracy had been dramatically decreased in FL~\cite{zizzo2020fat}. Furthermore, local data on clients are not necessarily or often not independent and identically distributed (IID)\cite{DBLP:journals/corr/abs-1806-00582}. The non-IID data make it even more challenging in training FL models so that FL has more difficulty in reaching a convergent aggregated model (also called {\it a global model}) \cite{zizzo2020fat, DBLP:books/sp/22/LB2022}. Moreover, non-IID data significantly reduce the robust accuracy of the global model. 

Addressing those challenges and combat the robust accuracy reduction, several studies proposed adversarial training~\cite{DBLP:journals/corr/GoodfellowSS14} and certifiable defense~\cite{DBLP:conf/mass/ChenKGZ21, DBLP:journals/corr/abs-2112-10525} in each client locally for FL. Zizzo et al.~\cite{zizzo2020fat} introduced federated adversarial training in 2020. As reported in Zizzo et al.~\cite{zizzo2020fat}, federated adversarial training is comparable to centralized adversarial training in the IID data case, but their obtained robust accuracies are less than 40\%. Furthermore, as shown in Shah et al.~\cite{shah2021adversarial}, FL reduces robust and natural accuracy compared to the centralized learning environment. Since adversarial training is highly dependent on hyperparameters, it is vital to study more advanced adversarial training methods for improving the robust and natural accuracy of FL.

In this paper, we propose an approach to studying ML robustness by simplifying flipping and cropping to generate a Gaussian noise in data argumentation and revising RestNet18 model architecture~\cite{He2015ResNet} in the centralized environment. In this research, evasion attack strategies, including Fast Gradient Sign Method (FGSM)~\cite{DBLP:journals/corr/GoodfellowSS14}, Carlini and Wagner (C\&W) attack \cite{DBLP:conf/sp/Carlini017}, DeepFool \cite{DBLP:conf/cvpr/Moosavi-Dezfooli16}, and Projected Gradient Descent (PGD) \cite{DBLP:journals/corr/MadryMSTV17}, are used to produce adversarial examples. Our experiments adopt Pytorch instead of the IBM toolbox used in  Lin, et al.~\cite{DBLP:journals/ejisec/LinNX22}. We apply soft labeling in adversarial training and add a Gaussian noise in training and testing to improve ML robustness. Our proposed centralized adversarial training approach achieves notable improvements in robust accuracy. Specifically, we increase the robust accuracy from 18.41\% to 65.4\% in FGSM attacks and from 47\% to 83.0\% in DeepFool attacks, respectively, by comparing the results in \cite{DBLP:journals/ejisec/LinNX22}. Moreover, we extend our approach to investigating ML robustness in FL. In the IID data case, our experimental results demonstrate that the resulting robust accuracy is comparable to the one obtained in the centralized environment. As noticed in Zizzo et al. \cite{zizzo2020fat}, there is a performance gap in ML robustness between IID and non-IID data. We propose the one- and two-classes approach for adversarial training in non-IID data to close up the performance gap. As noticed in our experiments, adversarial training with the one-class non-IID case results in poor performance in both natural and robust accuracy. We propose a data sharing approach for federated adversarial training in the one-class non-IID data
case. We achieve significantly higher natural
accuracy and robust accuracy than the non-data-sharing one.
In the two-class non-IID data case, the natural accuracy drops from 66.23\% to 57.82\%, and the robust accuracy decreases by 25\% and 23.4\% in C\&W and PGD, compared to the IID data case, respectively. We further propose an IID data sharing approach, which allows for increasing the natural accuracy to 85.04\% and the robust accuracy from 57\% to 72\% in C\&W and from 59\% to 67\% in PGD, respectively. This research has the following three major contributions.

\begin{itemize}
\item We propose an approach to simplifying data argumentation and modifying the ML model for understanding ML robustness in the centralized environment. We expand the study of ML robustness in Lin, et al.~\cite{DBLP:journals/ejisec/LinNX22} by considering the following major evasion attacks, including FGSM, PGD, C\&W, and DeepFool, under the white box attacks. Our experimental results demonstrate that the proposed approach achieves a significant improvement in ML robustness compared to Lin, et al.~\cite{DBLP:journals/ejisec/LinNX22}.

\item We extend the above approach in the centralized environment for investigating ML robustness in FL, where soft labeling is used in adversarial training and Gaussian noise is added in training and testing. We improve robust accuracy compared to Zizzo et al.~\cite{zizzo2020fat}.

\item We propose a one- and two-classes approach to studying ML robustness for IID and non-IID data in FL. Our experimental results demonstrate that the proposed approach can efficiently address the robust accuracy degradation of adversarial training in IID and non-IID data, which closes up the performance gap (reported in Zizzo et al.~\cite{zizzo2020fat}) between IID and non-IID data in FL.

\end{itemize}
The rest of the paper is organized as follows. Section~\ref{sec:background} first gives some background information related to this study and then presents the research problem with challenges. Section~\ref{sec:related_work} discusses related work. In Section~\ref{sec:method}, we give our proposed methodology. Section~\ref{sec:evaluation} discusses our experimental setups, datasets used in the research, and experimental results by comparing existing studies. Finally, we conclude our study with insights for future work in Section~\ref{sec:conclusion}.

\section{Background and Research Problem with Challenges} \label{sec:background}
We first give some background information about FL, adversarial attacks, and adversarial training, then present the research problem with challenges studied in this research.

\subsection{Background}
\subsubsection{Federated Learning (FL)}

\begin{figure}
    \centering
    \includegraphics[scale=0.9]{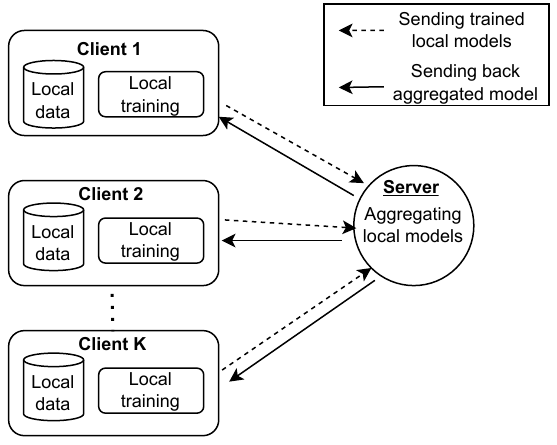}
    \caption{Model Training in FL}
    \label{fig:FL}
\end{figure}

As shown in Fig. \ref{fig:FL}, FL allows multiple devices (referred to as clients) to collaboratively train a model without sharing the local data of these devices~\cite{DBLP:books/sp/22/LB2022}~\cite{DBLP:conf/iclr/LiHYWZ20}~\cite{DBLP:journals/corr/abs-1902-04885}. The clients first train their models locally based on their own data and then send the trained models'hyperparameters to the FL server for aggregating model parameters. Once the model aggregation is finished, the server broadcasts an aggregated model (i.e., aggregated hyperparameters) to all the clients so that they can continue to train the models based on the updated model hyperparameters from the server by using their own local data. After the clients finish their model training locally, they will send their updated models to the server again for the model aggregation. This process will be repeated until a pre-defined criterion is met; A sample criterion could be a specified number of rounds or pre-defined accuracy. The resulting model is called a `global model'~\cite{DBLP:books/sp/22/LB2022}, which will be further used in the testing phase, and the performance of the global model will be evaluated on test data. There are multiple methods for model aggregation, such as FedAvg~\cite{DBLP:conf/iclr/LiHYWZ20}, FedCurv~\cite{DBLP:journals/corr/abs-1910-07796}, and FedProx~\cite{DBLP:conf/mlsys/LiSZSTS20}. For instance, FedAvg computes the weighted average of the model updates from all the clients~\cite{shah2021adversarial}. It must be pointed out that all data are kept privately among clients throughout the FL process. 
 
\subsubsection{Adversarial Attacks}
Szegedy et al. \cite{DBLP:journals/corr/SzegedyZSBEGF13, szegedy2016rethinking} first found that adversarial attacks would compromise neural network models as attackers can attempt to manipulate input data to make the neural networks untrustworthy. This issue limits the use of ML in real-world applications whose cyber and physical environments are not malicious free. Specifically, an adversarial attack refers to an attack such that an ML model misclassifies image data, while a human can easily classify them by eyes. Adversarial attacks can be targeted or untargeted. A targeted attack crafts an adversarial image $x^\prime$ to be classified as a specific class $t$, but the class, $t$, is not the actual class that $x$ belongs to. On the other hand, an untargeted adversarial attack does not target a specified class when the attack is generated. In this paper, we focus on untargeted attacks,  which are discussed as follows.


{a). \it Fast Gradient Sign Method (FGSM):}
 This method is introduced by Goodfellow et al. \cite{DBLP:journals/corr/GoodfellowSS14} for generating adversarial examples against deep learning models. The FGSM algorithm generates perturbations in a loss gradient direction using one single-step process~\cite{DBLP:journals/corr/abs-2112-02797}. Therefore, an attacker aims to increase the value of the loss function rather than decrease it by adding noise in the direction of the loss gradient, that is,
\begin{equation}
    \eta = \epsilon \times \sign (\nabla_{x}\mathcal{L}(f(x),y; \theta)),
\end{equation}
where $\epsilon$ is an imperceptible perturbation value that is predefined by the user, and $\sign (\nabla_{x}\mathcal{L}(f(x),y; \theta))$ represents the sign of the gradient of loss function $\mathcal{L}(f(x),y; \theta)$. Then, the adversarial image $x^\prime$ can be determined as follows:
\begin{equation}
    x^\prime = x + \eta
\end{equation}
{b). \it  Basic Iterative Method (BIM):}
Kurakin et al. \cite{DBLP:journals/corr/KurakinGB16} proposed an extended version of FGSM called BIM by applying FGSM multiple times. The initial point $x^{0}$ is set to the original point $x$, and then at each iteration, we update point $x^t$ by:
\begin{equation}
\label{PGD}
    x^{t+1} = Clip_{\epsilon}\left(x^t + \alpha \times \sign\nabla_{x}\mathcal{L}(f(x),y; \theta) \right),
\end{equation}
where $\alpha$ denotes the step size for each iteration, and the $Clip_{\epsilon}$ function truncates a perturbation larger than $\epsilon$ to make sure that the perturbation is within the $\epsilon$ bound; the $Clip_{\epsilon}$ function prevents a simple outlier detector from detecting a perturbed image. An adversarial example is obtained after a specified number of iterations, $m$. 

{c). \it Projected Gradient Descent (PGD):}
A PGD attack is another powerful variant of FGSM and BIM attacks. It applies a popular iterative approach PGD to maximize the loss~\cite{DBLP:journals/corr/MadryMSTV17}. PGD and BIM are similar except that PGD sets $x^0=x+\eta_{\epsilon}$, but BIM sets $x_0=x$, where $\eta_{\epsilon}$ is a random noise drawn from a uniform distribution in the range of $[-\epsilon, \epsilon]^n$, where $n$ is the dimension of $x$.

{d). \it Carlini and Wagner Attack (C\&W):}
The C\&W attack follows a different approach than the previously discussed attack methods. Carlini et al. \cite{DBLP:conf/sp/Carlini017} proposed to craft an adversarial example, $x'$, by solving the following optimization problem:
\begin{equation} \label{ep:C&W}
 \min \mathcal{D}(x,x')
\end{equation}
such that \[f(x')\leq 0\]
where $\mathcal{D}$ denotes the distance between original images $x$ and adversarial image $x'$, in a certain measure such as $L_0$ and $L_{\infty}$. $f(x')$ is an objective function such that $x'$ is misclassified if and only if $f(x')\leq 0$. Carlini et al. \cite{DBLP:conf/sp/Carlini017} provided a range of choices for objective function $f$ in their study. Based on our extensive experiments, C\&W attacks require a high computational cost compared to other attacks.

{e). \it DeepFool Attack:}
A DeepFool attack~\cite{DBLP:conf/cvpr/Moosavi-Dezfooli16} is initially used to attack binary classification problems and is then extended to muti-class classification problems.  To create an adversarial example, DeepFool finds the shortest distance to the decision boundary, where two classes are separated. In a binary classification problem, a DeepFool attack generates adversarial examples by solving the following optimization problem:
\begin{equation} \label{ep:deepfool}
 \min_{x'} \|x-x'\|_2
\end{equation}
such that \[f(x)+\nabla f(x)^T(x-x')=0,\]
where $\|x-x'\|_2$ denotes the $L_2$ distance between the original example and the perturbed image, and $f$ is the decision boundary. Moosavi-Dezfooli et al. \cite{DBLP:conf/cvpr/Moosavi-Dezfooli16} approximated $g$ using the first-order Taylor expansion and solved $x'$ iteratively to obtain an adversarial example.

\subsubsection{Adversarial Training}
In this subsection, we provide the necessary background about centralized and federated adversarial training for this study.

{a). \it Centralized adversarial training:}
Given $(x \in R^n, y \in k)$ and a loss function $\mathcal{L}$ for a neural network $f$ with model weights $\theta \in R^p$, where $k$ is the number of classes and $p$ is an integer representing the dimension of the model weights, we consider a classification task in ML. For this purpose, the main objective of the neural network training is to find $\theta$ that minimizes the loss function: 
\begin{equation} \label{ep:Train}
\theta^{*} = \argmin_{\theta} \EX_{(x, y) \sim D} [\mathcal{L}(f(x), y; \theta))], 
\end{equation}
where $D$ is a training set. The training process based on the above optimization problem is known as the empirical risk minimization (ERM). However, classifiers trained by this approach are vulnerable to adversarial examples.

To overcome the destruction caused by adversarial attacks, Madry et al. \cite{DBLP:journals/corr/MadryMSTV17} proposed an effective remedy called adversarial training. Specifically, the training process starts by feeding the perturbed inputs generated by PGD to the loss function $\mathcal{L}$ instead of training data. This gives rise to an adversarial training process. The goal of the adversarial process is to train on adversarial examples until the trained model can classify training data correctly. Hence, the adversarial training process is formulated as the composition of an inner maximization problem and an outer minimization optimization problem~\cite{DBLP:journals/corr/MadryMSTV17, shah2021adversarial}:
\begin{equation}
    \label{eq:AT}
    \min_{\theta}  \mathbb{E}_{(x, y) \sim D} \left[\max_{\|\delta\|_p \leq \epsilon} \mathcal{L}(f(x+\delta),y;\theta)\right],
\end{equation} 
where $\delta$ represents the perturbation added to the training input $x$, and $\epsilon > 0$ denotes the adversarial perturbation radius that is the maximum perturbation allowed.

{b). \it Federated Adversarial Training:}
FL is a distributed ML, where the local models with decentralized data are trained at edge devices (also referred to as {\it clients}) independently, and the resulting model parameters are aggregated in a server (also referred to as an {\it aggregator}) for further training. Specifically, suppose that there are $K$ clients, and each client has training data $D_{k}$ and model weights $\theta^{t}_{k}$ at the beginning of round $t$. The model weights among the clients are aggregated using a fusion function $F: \Rd^{K \times p} \rightarrow \Rd^p$ in the server. That is, each client solves the optimization problem given in Equation \ref{eq:AT} 
over their own datasets ($D_k)$ for local adversarial training, and then the server aggregates model weights after local adversarial training using fusion function $F$ for each round. Let $\theta_i^t$ is the adversarially trained model by the client $i$. Then, for a given round $t$, Equation \ref{ep:fl_fusion} formulates the aggregation of adversarially trained models by each client. 
\begin{equation} 
\label{ep:fl_fusion}
\theta^{t+1} = F(\theta_{1}^{t}, \theta_{2}^{t},...,\theta_{K}^{t}) 
\end{equation}


Since each device's local data is not shared with other devices, FL preserves each device's data privacy. Thus, Fl is applicable for various data sensitive applications.  
However, FL suffers from adversarial examples which direct ML models to incorrect classifications. Among evasion attack defense methods, adversarial training has become one of the most promising approaches. The research of this paper focuses on adversarial training for improving robust accuracy. 

\subsection{Research Problem with Challenges}
It is a known fact that current ML approaches, including centralized and decentralized, need to be more robust against evasion attacks. In this study, we focus on the adversarial attacks under white-box settings, where the adversary has complete access to our trained global classifier stored in the clients. Adversarial training has been one of the most popular remedies to address this issue under centralized environments. However, researchers find it challenging to extend the centralized methods for a decentralized environment due to several reasons, such as the difficulty of adapting to non-IID setting, higher performance degradation in both natural and robust accuracy compared to the centralized training, and the increase in the convergence time due to non-IID data~\cite{zizzo2020fat}. Therefore, existing adversarial training must be further developed so that ML can be successfully used in a decentralized environment, such as FL. We address the above challenges by improving existing adversarial training methods and extending them to FL with careful consideration of data distributions, i.e., IID and non-IID data.

\section{Related Work}\label{sec:related_work}
Lin, et al.~\cite{DBLP:journals/ejisec/LinNX22} proposed the adversarial retraining method for a centralized environment. It involves several strategies to improve the robustness of ML models. Label smoothing is one of these strategies. Label smoothing is an approach used in deep learning to reduce the tendency to overfit training data. This approach has been employed in several cutting-edge models, such as those used for image classification, language translation, and speech recognition~\cite{DBLP:conf/nips/MullerKH19}. 
To generate soft labels, Muller et al.~\cite{DBLP:conf/nips/MullerKH19} assigned a high probability (e.g., 0.9) to the correct class and then distributed the rest (e.g., 0.1) among other classes uniformly. Lin, et al.~\cite{DBLP:journals/ejisec/LinNX22} applied this soft labeling approach to adversarial training. However, they assigned the probabilities depending on the perturbation limit of adversarial examples. In contrast, we use a simpler approach by allocating the same probability scheme for all the adversarial examples in this paper. Moreover, Lin, et al.~\cite{DBLP:journals/ejisec/LinNX22} used adversarial examples generated by C\&W and PGD attacks for training, whereas we use only PGD examples to reduce the data generation time. Furthermore, we utilize a modified version of the ResNet-18 model in our experiments instead of the official ResNet18~\cite{He2015ResNet}. 

Zhou et al.~\cite{zhou2021adversarially} created a centralized environment based on an FL framework consisting of only one client (K = 1). We adopted the same technique to build our centralized experiments that can be readily expanded to the FL environment. Moreover, Zhou et al.~\cite{zhou2021adversarially} considered white-box attacks using FGSM and PGD. However, they obtained robust accuracy of less than 30\% in the centralized case. The centralized environment reported by Shah et al.~\cite{shah2021adversarial} achieved a 44\% robust accuracy, and Shah et al.~\cite{shah2021adversarial} obtained a comparable robust accuracy in the FL environment with IID data, where they considered five clients in their experiments. Zizzo et al.~\cite{zizzo2020fat} also reported robust accuracy of 38.58\% for the centralized case. Their IID FL setting with 51 clients offered comparable results to the centralized case. 

Compared to centralized adversarial training, decentralized adversarial training generally exhibits performance degradation, particularly in the scenarios where the training data are non-IID. For instance, studies by Zizzo et al. \cite{zizzo2020fat} and Shoham et al.~\cite{DBLP:journals/corr/abs-1910-07796} showed that a vanilla federated adversarial training model could handle IID data effectively. However, they also pointed out that the performance of models trained on non-IID data was significantly degraded compared to models trained on centralized and decentralized environments with IID data. Shoham et al.~\cite{DBLP:journals/corr/abs-1910-07796} focused on the feasibility of using adversarial training in federated learning with a fixed communication budget and non-IID data. 
To develop effective regularization techniques for federated adversarial training with non-IID data, Chen et al.~\cite{chen2022gear} introduced a class-wise cross entropy loss function, which forces the decision boundary for the minority classes to have larger margins. Thus, the global model can separate the unseen data of minority classes effectively. In addition to the challenge of non-IID data, Hong et al.~\cite{DBLP:journals/corr/abs-2106-10196} also addressed the issue of clients with low computational power. In their approach,  20\% of the clients are allowed to use adversarial training for learning, while the remaining clients with limited computational power can only afford standard training. Since the low resource clients lack adversarial examples during training, they proposed an effective sharing method called Federated Robust Batch-Normalization (FedRBN)~\cite{DBLP:journals/corr/abs-2106-10196} so that clients using the adversarial training method can share their adversarial robustness with other clients. Furthermore, it is worth noting that none of the previously proposed methods considered the modified ResNet-18 model architecture and dataset augmentation techniques employed in our method. As a result, these methods perform worse (lower natural accuracy and robust accuracy in the centralized training. 

\section{Methodology} \label{sec:method} 
\begin{figure}[!ht]
		         \centering
		         \includegraphics[scale=0.59]{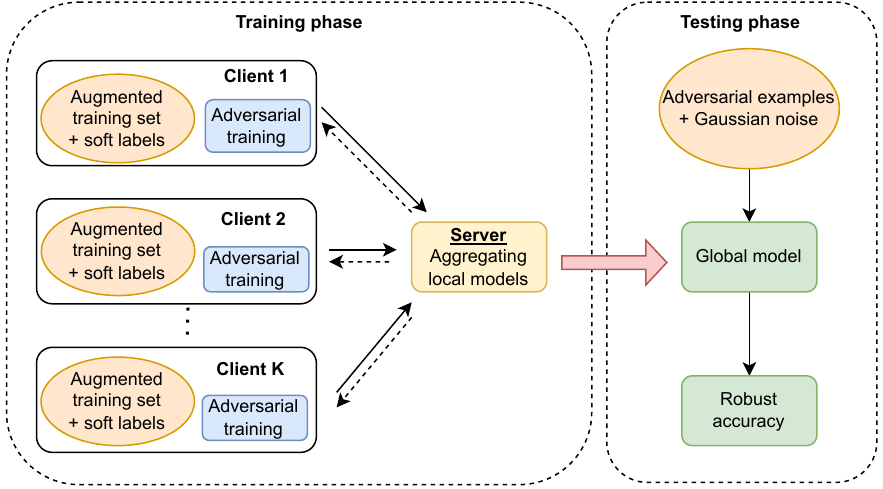}
		        \caption{Federated adversarial training procedure.}
		        \label{fig:process}
\end{figure}
\begin{figure}
         \centering
         \includegraphics[scale=0.5]{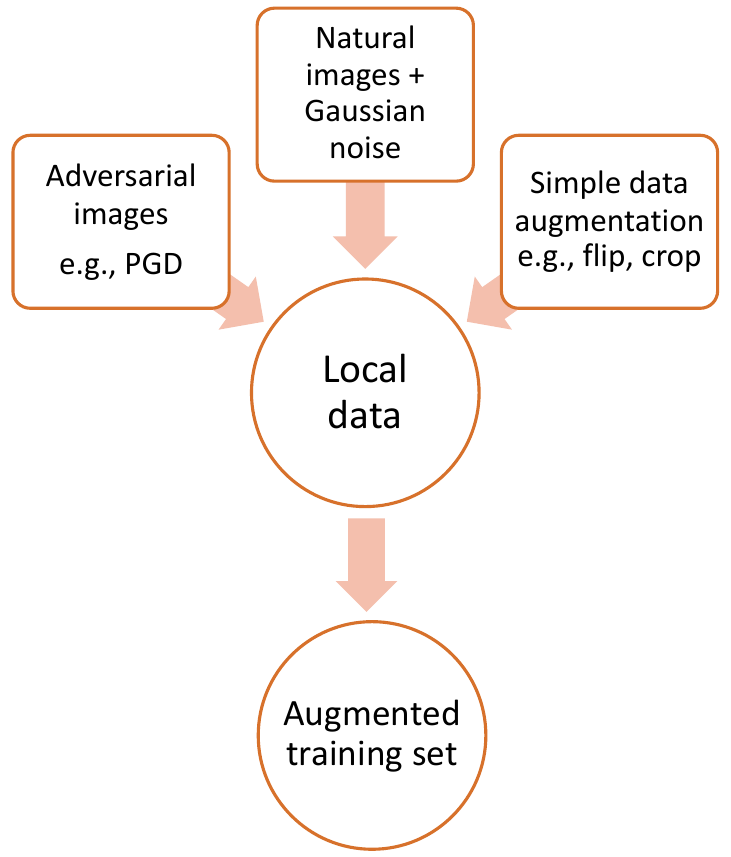}
        \caption{Augmenting training set for adversarial training by each client.}
        \label{fig:data_aug}
\end{figure}

We propose an iterative adversarial training process in Fig.~\ref{fig:process}. This figure summarizes the two phases of our proposed approach: 1) federated adversarial {\it training} and 2) adversarial example {\it testing} that evaluates the robustness of the global model. As the first step of the training phase, we augment the local data in each client, as shown in Fig.~\ref{fig:data_aug}. We expand the training set with adversarial examples crafted from PGD. We avoid generating adversarial images with high perturbations, as human eyes easily detected such images. We aim to develop an approach to defend against adversarial images that are unable to detect through human eyes. Additional examples are generated by adding Gaussian noise to normal or natural images in order to make our model robust against random noise. Furthermore, simple data processing methods like horizontal flipping and random cropping with padding are also applied to the training set to reduce overfitting \cite{DBLP:journals/jbd/ShortenK19}. 
Next, we incorporate soft labeling for the target values instead of hard labeling using the approach mentioned in Muller et al. \cite{DBLP:conf/nips/MullerKH19}. Instead of assigning a probability of `1' to the true label and `0' to the rest of the labels, we assign a high probability closer to 1 to the true label and a very low probability to the rest of the labels while the sum of all the probabilities is still 1. Soft labeling is widely used to avoid the overfitting of the models and improve generalization\cite{DBLP:conf/nips/MullerKH19}. Equation \ref{eq:soft_label} defines soft labels for each input. Consider an input that belongs to class $c$, where $c$ is an integer between 1 and the total number of classes $N$. Let $\mathbf{y}$ denote its ground truth label with the $i$-th element of $\mathbf{y}$
\begin{equation}
    y_i= \begin{cases}
        1 & \text{if } i =c\\
        0 & \text{if } i \neq c,
    \end{cases}
\end{equation}
for $i=1, \dots, N$. Then, the soft label $\mathbf{y}^{SL}$,
\begin{equation}
\label{eq:soft_label}
    y_i^{SL}= \begin{cases}
       1- \frac{N-1}{N}\alpha & \text{if } i =c\\
         \frac{\alpha}{N} & \text{if } i \neq c,
    \end{cases}
\end{equation}
where $\alpha$ is a label smoothing parameter that is close to 0. Using the soft labels and the augmented training examples, each client iteratively trains the model assigned by the server for a given number of iterations.

In this study, we used the Federated Average algorithm (FedAvg) \cite{DBLP:conf/aistats/McMahanMRHA17} as the model aggregation method. At the end of communication round $t$ between clients and server, the server collects the local model parameters $\theta_{k}^{t}$ from each worker $k$. To aggregate local models, the local model weights are averaged based on their data contribution ratio $\frac{|D_k|}{|D|}$, as shown in the following equation
\begin{equation}\label{fedavg}
    \theta^{t+1} = \frac{1}{|D|} \sum_{k=1}^K{|D_k| \theta_{k}^{t}},
\end{equation}
where $|D_k|$ is the size of the dataset belongs to the client $k$ and $|D|=\sum_{k=1}^K |D_k|$.
Next, we discuss the our federated adversarial training framework, which is based on Federated Averaging Adversarial Training (FedAvgAT) algorithm proposed in Shah et al. \cite{shah2021adversarial}. FedAvgAT is an extension of FedAvg algorithm in the adversarial federated environment, which considers adversarial examples when generating the local model updates. Algorithm \ref{algo:fl} provides a detailed representation of the algorithm employed in our federated adversarial training approach. We describe the main components of Algorithm \ref{algo:fl} as follows.  

\begin{algorithm}
\begin{algorithmic}[1]
\STATE \textbf{Input} $K$: number of clients; $R$: number of communication rounds; $A_T$: augmented training set; $y$: the training set labels; $\alpha$: a label smoothing parameter; $E$: number of local training epochs per round; $\theta^{0}$: initial model weights.
\STATE \textbf{Output}: The global model that is robust to adversarial attacks. 
    \FORALL{ $t = 1$ \textbf{to} $R$}
        \STATE Aggregator sends the model weight $\theta^{t}$ to all clients
        \FORALL{$k = 1$ \textbf{to} $K$}
            \STATE Calculate $\mathbf{y}_{SL}$ based on Equation~\ref{eq:soft_label}
            \STATE $\theta_{k}^{t+1} = AT(E, \theta^{t}, A_{T}, \mathbf{y}_{SL})$; 
            \STATE Send $\theta_{k}^{t+1}$ back to the aggregator;
        \ENDFOR
        \STATE $\theta^{t+1} = FedAvg(\theta_{1}^{t}, \theta_{2}^{t},...,\theta_{K}^{t})$; 
    \ENDFOR
\end{algorithmic}
\caption{Federated adversarial training with soft labeling} 
\label{algo:fl}
\end{algorithm}
First, each client trains the model using their private data for $E$ epochs locally. Then, each client sends the weights of their adversarially trained model $\theta_{k}^{t}$ to the server. The server aggregates these weights using a fusion function $F$, such as FedAvg~\cite{DBLP:conf/aistats/McMahanMRHA17}, and sends back the updated weights $\theta^{t+1}$ to the clients. This completes one communication round between the server and clients. The process is repeated for $R$ communication rounds before obtaining the final global model.

Gaussian noise is added to the test images when testing the global model robustness. Since the model is trained with Gaussian noise, the model is robust against the small random noise. In contrast, when the noise is added to adversarial images, it will likely distort their deliberately calculated adversarial perturbations~\cite{DBLP:journals/ejisec/LinNX22, DBLP:conf/icc/LinNX20}.

As part of this study, we also consider the following two cases: 1) clients have IID data, and 2) clients do not have IID (non-IID) data. In the IID case, each client is expected to have the same amount of data with the same class distribution. In the non-IID case, clients will have a different amount of data with non-uniform class distributions. To create non-IID data among the clients, we used two approaches described in Zhao et al. \cite{DBLP:journals/corr/abs-1806-00582}. In the first approach, we assigned data from only one class to each client. Hence, the class distribution among the clients is highly deviated from the IID case. 
In the second approach, we randomly assigned data from two classes to each client.
We further apply the strategy proposed by Zhao et al. \cite{DBLP:journals/corr/abs-1806-00582} that involves data sharing to mitigate the skewness of the non-IID data. 

\section{Experimental Evaluation and Results} \label{sec:evaluation}
As we discuss later, we will present the implementation and evaluation of the proposed adversarial (re)-training system with experimental results. We evaluate the effectiveness of our centralized adversarial training approach and federated adversarial training approach with both IID and non-IID data. We aim to demonstrate the efficacy of both approaches by comparing natural and robust accuracy and by closing the performance gap between the centralized and federated adversarial training.

\subsection{Software and Hardware}
In contrast to our study, which used Python 3.7.6 and PyTorch 1.13.1 to generate adversarial examples for training and testing, Lin, et al.~\cite{DBLP:journals/ejisec/LinNX22} adopted the Adversarial Robustness Toolbox (ART)~\cite{art2018}. While our approach allowed for fine-grained control over the generation of adversarial examples, the use of ART is provided in the study~\cite{DBLP:journals/ejisec/LinNX22} with a more standardized and automated process for generating and evaluating the (re)-trained models. Despite these differences in implementation, both studies aim to evaluate the model\textcolor{blue}{'}s robustness. It is worth mentioning that our implementation of FL is a simulated version, as we created multiple variables within a Python program to represent a central server and multiple client nodes. Therefore, there was no actual communication between the server and its clients. To accelerate computations, we use a shared NVIDIA A40 GPU card to reduce the training time of a model. However, when A40 is occupied by other university researchers, we use a NVIDIA GeForce GTX 1080 Ti card.
\subsection{Data and Models}
\begin{figure*}
    \centering
    \includegraphics[scale=0.5]{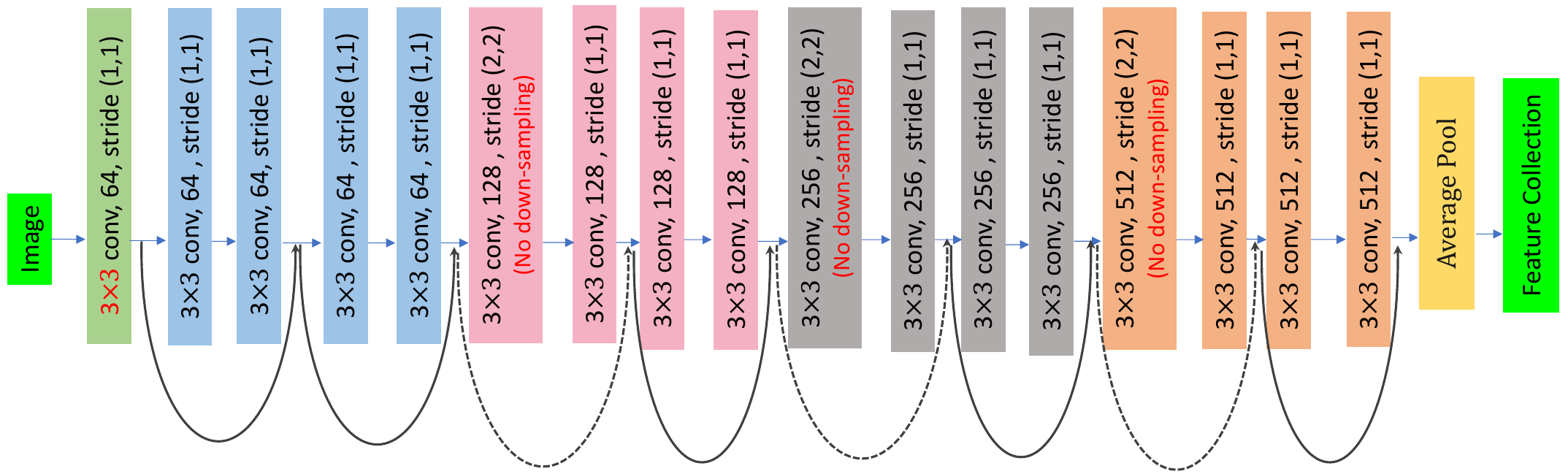}
    \caption{The modified ResNet18 model, where its first convolutional layer has a kernel of size $3\times3$ instead of $7\times7$, and the modified model removes down-samplings (red) in ResNet18~\cite{He2015ResNet}.}
    \label{fig:model_architecture}
\end{figure*}

For our study, we evaluate the effectiveness of the proposed methods using CIFAR-10 dataset~\cite{krizhevsky2009learning}, which is widely used as a benchmark in adversarial machine learning research. For experiments with the CIFAR-10 dataset, we use the ResNet-18 architecture~\cite{He2015ResNet}. However, we make two modifications, as shown in Fig.~\ref{fig:model_architecture}. First, we change the kernel size of the first convolution layer. Secondly, we remove the down-sampling part in all ResNet blocks except the first block, which does not have the down-sampling. We trained the modified ResNet-18 architecture from scratch for all experiments conducted in this study. 
As a result, the modified model has 11,173,962 learnable parameters, which is fewer than the 11,689,512 parameters in the official ResNet-18 architecture~\cite{He2015ResNet}.

\subsection{Centralized Adversarial Training} \label{sec:centralized_AT}
To evaluate the robustness of our proposed centralized model on the CIFAR-10 dataset, we adopted the setup described in Lin, et al.~\cite{Li2020}. We trained the model with PGD-based adversarial examples by setting the perturbation magnitude in each iteration $\epsilon = 8/255$, the number of iterations $n_{iter} = 7$, and the step size $\alpha = 2/255$. Additionally, we augmented the training set with Gaussian noise with a mean of 0 and a standard deviation of 0.1. Details about how to generate the adversarial images were discussed in the Methodology section (section \ref{sec:method}). We adversarially trained the centralized model using the SGD optimizer with a learning rate of 0.001, a momentum rate of 0.9, and a weight decay rate of 0.0002. The learning rate was adjusted during training using a function that reduced the initial learning rate by a factor of ten after 100 and 150 epochs, respectively. Specifically, the initial learning rate is 0.1. If the current epoch is greater than or equal to 100, the learning rate is divided by ten. If it is greater than or equal to 150, the learning rate is divided by ten again. We chose categorical cross entropy as our loss function. At test time, we evaluate the robustness of our model against adversarial attacks, including FGSM, C\&W, DeepFool, and PGD. Before testing, we added Gaussian noise with a mean of 0 and a standard deviation of 0.1 to the test images. The experiment results are presented in Table \ref{table:centralized_AT_robustness}, where we compare our method with Lin, et al.'s approach \cite{DBLP:journals/ejisec/LinNX22}.

\subsection{Federated Adversarial Training with IID and Non-IID Data}
In the subsequent sections, we will discuss how to prepare for both IID and non-IID data in the context of our proposed decentralized adversarial training method. We conducted all our experiments with $K$ workers, $K \in \{5, 10\}$. When generating adversarial examples, we use the PGD algorithm and Gaussian noise with the same hyperparameters as those used in the centralized adversarial training case (section ~\ref{sec:centralized_AT}). The number of local epochs we experiment with is $E=1, 3, 5$.

\subsubsection{IID Data}
In the IID CIFAR-10 data split, the training set is randomly partitioned into disjoint subsets, with each subset following a uniform data distribution across the ten classes of the CIFAR-10 dataset. These subsets are assigned to individual clients, with the number of subsets equal to the number of clients. The data partition ensures that each client's training set follows an IID data distribution, an essential assumption in training machine learning models. Each client then trains its local model on its private data. In Section \ref{FAT_IID}, we will compare the results of our federated adversarial learning with IID data when the number of clients is five and ten.


\subsubsection{Non-IID Data} \label{Non-IID_dataset}
In contrast to the IID data split, where the training set is randomly partitioned into disjoint subsets with each subset following a uniform data distribution, non-IID data split involves partitioning the data such that each subset has a unique data distribution. This method results in each client having a non-uniform subset of the training data. To create a non-IID training subset, the training sets are distributed to each client in a highly heterogeneous manner such that each client is randomly allocated either one or two classes of data, as also done in \cite{DBLP:journals/corr/abs-1806-00582}. It makes all clients' data distribution different from each other and skewed towards the assigned subset of classes. The details of our analysis and the impact of non-IID data on our method were presented in Section \ref{FAT_Non_IID}.

\begin{table*}[ht!]
    \begin{tabular}{|p{3.4cm}| p{3cm}| p{1.6cm}|p{2.4cm} |p{2.4cm}| p{2.3cm}|}
    \hline
    Training Set & Model & Learning Rate &Test Set & \textbf{Natural Accuracy} & \textbf{Robust Accuracy} \\
    \hline 
    Natural examples & Customized ResNet18 & Varied & FGSM & 99.26\% & 3.37\% \\
    \hline
    PGD examples & Customized ResNet18 & Varied & FGSM & 78.17\% & 47.05\% \\
    \hline
    FGSM examples & Customized ResNet18 & Varied & FGSM & 78.59\% & 27.27\% \\
    \hline
    PGD examples + Gaussian & Customized ResNet18 & Varied & FGSM + Gaussian & 78.65\% & 65.41\% \\
    \hline
    PGD examples + Gaussian & Official ResNet18 & Varied & FGSM + Gaussian & 70.17\% & 44.82\% \\
    \hline
    PGD examples + Gaussian & Customized ResNet18 & Fixed & FGSM + Gaussian & 79.87\% & 55.95\% \\
\hline
\end{tabular}
\caption{Centralized adversarial training - Natural and robust accuracy (\%) on the CIFAR-10 test set. Adversarial examples based on FGSM  with $\epsilon = {8}/{255}$ are used to measure robust accuracy. }
\label{table:centralized_AT}
\end{table*}

\subsubsection{Local Adversarial Training} \label{Local_training}
We adapt our centralized adversarial training method to each client to countermeasure adversarial attacks. We use the same set of hyperparameters for crafting adversarial examples and training local models across all clients as thosed used in the centralized training case (section \ref{sec:centralized_AT}). Our experimental setup assumes that all clients are selected to participate in every communication round, which is consistent with the approach taken by Shah et al.~\cite{shah2021adversarial}. 
\subsection{Experimental Results}
In this section, we present the experimental results of our proposed model's performance on natural and adversarial examples under both centralized and federated adversarial training scenarios. Our evaluation includes assessing the model's generalization ability and robustness against adversarial attacks at test time. Our method achieves notable improvements in centralized training and federated adversarial training with IID and non-IID data.
 
\subsubsection{Centralized Training} \label{Centralized_training_result}
The ResNet-18 trained on CIFAR-10 achieves a normal accuracy of 99.26\% and a robust accuracy of 3.37\% on CIFAR-10 test set. When we do centralized adversarial training using PGD examples, the natural and robust accuracy of the customized ResNet-18 model is 78.17\% and 47.05\%, respectively. When the training dataset is also augmented with Gaussian noise examples, the robust accuracy (65.41\%) improves significantly. In contrast, if replacing PGD adversarial examples with FGSM examples with $\epsilon = {8}/{255}$, the robust accuracy reduces from 47.05\% to 27.27\%. This shows that PGD examples, generated via an iterative local search process around the neighborhood of the original examples, are superior. Additionally, compared to the official ResNet-18~\cite{He2015ResNet}, our customized ResNet-18 achieved superior performance in both natural accuracy (78.65\% vs 70.17\%) and robust accuracy (65.41\% vs 44.82\%). If we fix the learning rate to 0.1 instead of varying it, there is a slight increase in natural accuracy (from 78.17\% to 79.87\%) but a dramatic decrease in robust accuracy (from 65.41\% to 55.95\%). The key result is summarized in Table \ref{table:centralized_AT}. Hence, we use the customized ResNet-18 model trained with varying learning rates as the candidate for comparison with Lin, et al.~\cite{DBLP:journals/ejisec/LinNX22} method in Table \ref{table:centralized_AT_robustness}.

Table \ref{table:centralized_AT_robustness} illustrates the improvement over Lin, et al.`s method~\cite{DBLP:journals/ejisec/LinNX22} and highlights the effectiveness of our centralized adversarial training method in achieving robust accuracy against different white-box attacks. Our method achieves better robust accuracy against the FGSM, C\&W, and DeepFool attacks. 

\begin{table}[H]
    \centering
    \begin{tabular}{|l|l|l|l|l|l|l|}
    \hline
        Training & FGSM & C\&W & DeepFool \\ \hline
        Lin, et al.~\cite{DBLP:journals/ejisec/LinNX22} & 47\% & 78\% & 36\%  \\ \hline
        Our method & 65.41\% & 81\% & 83\%  \\ \hline
    \end{tabular}
    \caption{Centralized adversarial training - Robust accuracy (\%) on the CIFAR-10 test dataset under various white-box attacks.}
    \label{table:centralized_AT_robustness}
\end{table}

\subsubsection{Federated Adversarial Training with IID Data} \label{FAT_IID}
In this subsection, we compare the performance of federated adversarial training with IID data to the centralized adversarial training case. Our aim is to demonstrate that the federated approach can achieve comparable robustness against adversarial attacks. As shown in Table~\ref{table:FAT_AT-IID}, in the federated case with five clients, the robust accuracy for C\&W is similar to the centralized model, whereas for DeepFool attacks, it is only within 4\% of the centralized model. For the PGD attack, the federated approach with five clients is slightly worse than the centralized case. We observe a similar trend with ten clients. In detail, the federated approach's performance is within 5\% and 7\% of the centralized adversarial training results for the C\&W and DeepFool attacks, respectively. For the PGD attack, the federated approach with ten clients is 5\% better than the centralized case. Based on our experimental results, it is evident that federated adversarial training with IID data can yield comparable robustness to the centralized case, especially when all participants are included in all communication rounds.
\begin{table}[!ht]
    \centering
    \begin{tabular}{|l|l|l|l|l|l|l|}
    \hline
        \# Clients & Natural & FGSM & C\&W & DeepFool & PGD\\ \hline
        K = 5 & 80.76\%  & 63.07\%  & 81\% & 79\% & 71\% \\ \hline
        K = 10 & 66.23\%  & 51.51\%  & 76\% & 76 \% & 77\% \\ \hline
    \end{tabular}
    \caption{IID Federated adversarial training - Robust accuracy (\%) on the CIFAR-10 test dataset under various white-box attacks. (Note: $R = 100$ and $E = 3$).}
    \label{table:FAT_AT-IID}
\end{table}
\subsubsection{Federated Adversarial Training with Non-IID Data}
\label{FAT_Non_IID} 
In this section, we explore the impact of data heterogeneity on the performance of our federated adversarial training method. Specifically, we investigate the effect of one-class and two-class non-IID data on both the natural accuracy and the robust accuracy of the global model. We also examine the effectiveness of the data sharing technique \cite{DBLP:journals/corr/abs-1806-00582} that allows clients to contribute a small portion of their private data to mitigate the impact of data heterogeneity. The experimental results are presented in Table~\ref{table:oneclass_non_iid_ed_AT_robustness_r100} and Table~\ref{table:twoclass_non_iid_ed_AT_robustness_r100}. In these tables, we compare the robust accuracy of our federated adversarial training method on the CIFAR-10 test dataset with and  without using the data sharing method for the local training\cite{DBLP:journals/corr/abs-1806-00582}. 

To create the global shared training set on CIFAR-10, we adopt the approach described in \cite{DBLP:journals/corr/abs-1806-00582}. We randomly choose 1,000 images per class from the training images, resulting in 10,000 images set aside for sharing. In our experiments, we randomly sample 500 images per class from the global shared dataset to reduce the impact of data heterogeneity. This leaves the remaining 5,000 images unused. After creating the global training subset, we divide the unselected training examples into multiple partitions. Each partition is assigned to a client for local adversarial training. The details of generating each client's non-IID data have been described in Section \ref{Non-IID_dataset}. After generating the global shared training set, we craft the non-IID data for each client as described in Section \ref{Non-IID_dataset}.

Our experimental results show that the federated adversarial training framework without data sharing achieves lower robust accuracy than the federated adversarial training framework with data sharing in both one-class and two-class non-IID data. Specifically, in the one class non-IID dataset, which is a more challenging and severe case of data heterogeneity than the two-class non-IID data, the global model trained without using the global shared training subset achieves low performance in both natural accuracy and robust accuracy. Specifically, the natural accuracy is 10.97\% and the robust accuracy is 1.61\%, 11\%, 12\%, and 11\% against FGSM, C\&W, DeepFool, and PGD attacks, respectively. On the other hand, the one-class non-IID federated adversarial training method with data sharing achieved a significantly higher natural accuracy and robust accuracy compared to the non-data-sharing case, shown in Table~\ref{table:oneclass_non_iid_ed_AT_robustness_r100}. In the two-class non-IID case, similar trends were observed, e.g., data sharing achieves significantly higher natural and robust accuracy compared to the non-data-sharing version. More detailed results can be found in Table \ref{table:twoclass_non_iid_ed_AT_robustness_r100}.
The result suggests that contributing a small portion of private data can help reduce the impact of data heterogeneity and recover the robustness of the federated learning model against the adversarial attacks.

\begin{table*}[!h]
    \centering
    \begin{tabular}{|p{7cm}|p{1.5cm}|p{1.5cm}|p{1.5cm}|p{1.5cm}|p{1.5cm}|p{1.5cm}|}
    \hline
        Training & Natural & FGSM & C\&W & DeepFool & PGD\\ \hline
        One class non-IID federated AT \textbf{without} data sharing & 10.97\%  & 1.61\%  & 11\% & 12\% & 11\% \\ \hline
        One class non-IID federated AT \textbf{with} data sharing  & 67.42\%  & 41.18\%  & 53\% & 47\% & 48\% \\ \hline
    \end{tabular}
    \caption{One class non-IID Federated adversarial training (AT) - Robust accuracy (\%) on the CIFAR-10 test dataset under various white-box attacks, where K = 10, R = 100, and E = 1.}
    \label{table:oneclass_non_iid_ed_AT_robustness_r100}
\end{table*}

\begin{table*}[!ht]
    \centering
    \renewcommand{\arraystretch}{1}
    \begin{tabular}{|p{7cm}|p{1.5cm}|p{1.5cm}|p{1.5cm}|p{1.5cm}|p{1.5cm}|p{1.5cm}|}
    \hline
        Training & Natural & FGSM & C\&W & DeepFool & PGD\\ \hline
        Two-class non-IID federated AT \textbf{without} data sharing & 
        57.82\%  & 54\%  & 57\% & 62\% & 59\% \\ \hline
        Two-class non-IID federated AT \textbf{with} data sharing  & 85.04\%  & 63.97\%  & 72\% & 71\% & 67\% \\ \hline
    \end{tabular}
    \caption{Two-class non IID Federated adversarial training (AT) - Robust accuracy (\%) on the CIFAR-10 test dataset under various white-box attacks, where K = 10, R = 100, and E = 3.}
    \label{table:twoclass_non_iid_ed_AT_robustness_r100}
\end{table*}

\section{Conclusion and Future Work} 
\label{sec:conclusion}
Although FL can offer significant privacy advantages, the reliability of FL models can be put into question by the existence of adversarial examples at the inference stage. In this study, we investigated ML robustness via adversarial training in both centralized and decentralized environments. We extended the adversarial training approach proposed by Lin, et al.~\cite{DBLP:journals/ejisec/LinNX22} and enhanced the ML robustness under a centralized environment using our proposed approach. In a decentralized environment, we investigated FL robustness for IID and non-IID data. Finally, we demonstrated that sharing some IID training data~\cite{DBLP:journals/corr/abs-1806-00582} can effectively mitigate the negative impact of non-IID data in FL, leading to improved robustness. Overall, our study suggests that the proposed method can effectively enhance the robustness of ML models against adversarial attacks in centralized and decentralized training settings. 

In the future, we expect to explore the ML robustness for other datasets, such as MNIST~\cite{DBLP:journals/pieee/LeCunBBH98}, ImageNet~\cite{russakovsky2015imagenet}, and Fashion-MNIST~\cite{DBLP:journals/corr/abs-1708-07747}. Furthermore, we intend to examine how ML robustness performs under decentralized environments when more clients (more than 10) exist and consider more non-IID data cases in the future. Lastly, Nasr et al.~\cite{DBLP:conf/sp/NasrSH19} showed a high amount of information leakage from local training data caused by inference attacks, while the global model achieves high accuracy. Therefore, we will consider studying the resilience of federated adversarial training against information leakage attacks in the future.

\section*{Acknowledgment}
We acknowledge the National Science Foundation (NSF) for partially sponsoring the work under grants \#1620871, \#1620862, \#1633978, \#1620868, \#2228562, and \#223683, as well as BBN/GPO project \#1936 through an NSF/CNS grant. We also thank the Florida Center for Cybersecurity (Cyber Florida) for a seed grant. The views and conclusions contained herein are those of the authors and should not be interpreted as necessarily representing the official policies, either expressed or implied by NSF.
\bibliographystyle{IEEEtran}
\bibliography{refs.bib}
\end{document}